\documentclass[11pt]{article}

\usepackage[preprint]{acl}

\usepackage{times}
\usepackage{latexsym}
\usepackage{amsmath}
\usepackage{xcolor}

\usepackage[T1]{fontenc}

\usepackage[utf8]{inputenc}

\usepackage{microtype}

\usepackage{inconsolata}

\usepackage{graphicx}
\usepackage{tikz}
\usetikzlibrary{arrows.meta}

\usepackage{amssymb}
\usepackage[table]{xcolor}
\usepackage{colortbl}
\usepackage{graphicx}
\usepackage[textsize=tiny]{todonotes}
\usepackage{diagbox}
\usepackage{booktabs} %
\usepackage{tablefootnote}
\usepackage{enumitem}
\usepackage{array}
\usepackage{makecell}
\usepackage{multirow}
\usepackage{relsize}

\usepackage{placeins} %
\usepackage{float} %
\usepackage{listings} %

\usepackage{tabularx}

\NewDocumentCommand{\coltxt}{O{gray}+m}{%
  {\color{#1}#2}%
}
\NewDocumentEnvironment{ColTxt}{O{gray}}{%
  \begingroup
  \color{#1}%
}{%
  \endgroup
}
\newcommand{\entarg}[1]{\ensuremath{\text{\relsize{-1}\bfseries #1}}}

\newcommand{\entails}[2]{%
  \ensuremath{%
    \text{\relsize{-1}\bfseries #1}%
    \mkern3mu\mathord{\sqsubseteq}\mkern3mu%
    \text{\relsize{-1}\bfseries #2}%
  }%
}
\newcommand{\entailssame}[2]{%
  \ensuremath{%
    \text{\relsize{0}\bfseries #1}%
    \mkern3mu\mathord{\sqsubseteq}\mkern3mu%
    \text{\relsize{0}\bfseries #2}%
  }%
}
\newcommand{\entailsnormal}[2]{%
  \ensuremath{%
    \text{\bfseries #1}%
    \mkern3mu\mathord{\sqsubseteq}\mkern3mu%
    \text{\bfseries #2}%
  }%
}

\newcommand{\baseoracle}{{\scriptsize$\blacksquare$}}
\newcommand{\LEX}{\textsc{lex}}
\newcommand{\LEXs}{\textsc{lex}s}

\usepackage[most]{tcolorbox}
\usepackage{enumitem}

\newtcolorbox{promptbox}[1][]{
    enhanced,
    breakable,
    colback=gray!3,
    colframe=gray!55,
    boxrule=0.6pt,
    arc=1mm,
    left=2mm,
    right=2mm,
    top=1.5mm,
    bottom=1.5mm,
    fontupper=\small,
    title={#1},
    fonttitle=\small\bfseries,
    coltitle=black,
    colbacktitle=gray!15,
    attach boxed title to top left={xshift=2mm,yshift=-1mm},
    boxed title style={
        boxrule=0.4pt,
        colframe=gray!55,
        arc=1mm
    }
}

\newcommand{\promptinput}[2]{%
    \textbf{Premise:} \textit{#1}\par
    \textbf{Hypothesis:} \textit{#2}\par
}

\newcommand{\promptanswer}[1]{%
    \textbf{Answer:} \texttt{#1}\par
}

\newcommand{\promptrelations}[1]{%
    \textbf{Relations:} #1\par
}

\newtcbox{\lexboxraw}[1][]{%
  enhanced,
  colback=white,
  colframe=black,
  boxrule=0.4pt,
  arc=2mm,
  left=2pt,
  right=2pt,
  top=2pt,
  bottom=2pt,
  boxsep=0pt,
  nobeforeafter,
  #1
}

\newcommand{\lexbox}[2][]{%
  \lexboxraw[#1]{%
    \begin{tabular}{@{}l@{}}#2\end{tabular}%
  }%
}
\tcbset{
  dottedlex/.style={
    boxrule=0pt,
    frame hidden,
    borderline={0.6pt}{0pt}{black,dashed}
  }
}
\newcommand{\lexempty}{\lexbox{$\varnothing$}}

\newtcbox{\wndeltabox}{
  on line,
  colback=red!5,
  colframe=red!45!black,
  boxrule=0.2pt,
  arc=0pt,
  boxsep=0pt,
  left=1.5pt,
  right=1.5pt,
  top=1pt,
  bottom=1pt,
  fontupper=\footnotesize
}

\newtcbox{\agentdeltabox}{
  on line,
  colback=blue!8,
  colframe=blue!45!black,
  boxrule=0.2pt,
  arc=1mm,
  boxsep=0pt,
  left=1.5pt,
  right=1.5pt,
  top=1pt,
  bottom=1pt,
  fontupper=\footnotesize
}

\newcommand{\dWN}[1]{\wndeltabox{#1}}
\newcommand{\dA}[1]{\agentdeltabox{#1}}
\newcommand{\dgap}{\hspace{0.55em}}

\newcommand{\dWNhead}{\wndeltabox{$\Delta_{\text{\textsc{wn}}}$}}
\newcommand{\dAhead}{\agentdeltabox{$\Delta_a$}}

\newcommand{\LLMa}{\ensuremath{\mathrm{LLM}_{a}}}
\newcommand{\LLMWN}{\ensuremath{\mathrm{LLM}_1^{\text{\!\textsc{wn}}}}}
\newcommand{\LLMaWN}{\ensuremath{\mathrm{LLM}_{a}^{\!\text{\textsc{wn}}}}}
\newcommand{\LLMone}{\ensuremath{\mathrm{LLM}_{1}}}

\title{Explaining Textual Entailment with Lexical Entailments:\\
Using LLMs to Supply Lexical Relations for Formal Proofs}

\author{
 \textbf{Jorryt de Jong\textsuperscript{1}} ~~~
 \textbf{Stefan Morača\textsuperscript{1}} ~~~
 \textbf{Ettore Cesari\textsuperscript{1}} ~~~
 \textbf{ Lasha Abzianidze\textsuperscript{2}}
\\
 \textsuperscript{1}Utrecht University, the Netherlands
\\
 \textsuperscript{2}Institute for Language Sciences, Utrecht University, the Netherlands
\\
 \texttt{\{j.e.j.dejong,s.moraca,e.cesari\}@students.uu.nl}
\\
\small{\textbf{Correspondence:} \href{mailto:l.abzianidze@uu.nl}{l.abzianidze@uu.nl}}
}

\begin{document}
\maketitle
\begin{abstract}
Large Language Models (LLMs) are highly capable of natural language reasoning and appear to store a great deal of lexical knowledge, but it is still unclear how much of this knowledge they actually use when reasoning, and whether they use it in the right way.
On the other hand, logic-based Natural Language Inference (NLI) systems provide transparent and formally grounded reasoning, but they need to be supplied with rich lexical knowledge to prove inferences beyond purely logical ones.
In this paper, we evaluate whether LLMs can identify all lexical knowledge needed to solve NLI problems and how much this knowledge contributes to proof search in a logic-based NLI system.
Our research focuses exclusively on structured lexical entailments (e.g., \entailssame{chinchilla}{small animal}) as a proxy for structured explanations for NLI problems with an entailment label. 
First, we curate a dataset for a new task of explaining sentential entailments with a set of lexical entailments.
The dataset is used to intrinsically evaluate LLMs on generating structured lexical explanations.
Then, we use NLI as an extrinsic evaluation in a simple neuro-symbolic setting, assessing whether LLMs can supply sufficient lexical relations to LangPro, a natural-logic theorem prover for natural language.
The results show that the proposed task remains challenging even for hosted proprietary LLMs, and that their contribution to theorem proving is moderate: generated relations are often only partially sound and may be tailored to the specific NLI problem rather than representing generally valid lexical knowledge.
\end{abstract}

\section{Introduction}

Recent LLM progress has shifted much of NLP research toward downstream tasks.
However, Natural Language Inference (NLI, \citealp{rteBook:2013}), a generic reasoning task, remains popular because it allows evaluation of models on specific reasoning phenomena.
The NLI task challenges models to predict whether a hypothesis sentence is entailed by, contradicted by, or is neutral with respect to a premise text.
While the frontier models have saturated common NLI benchmarks, we still do not know how sound their reasoning is when arriving at correct inference labels.

\begin{figure}[!t]
    \centering
    \includegraphics[width=.48\textwidth, 
    keepaspectratio,
    trim=2mm 3mm 170mm 1mm,
    clip]{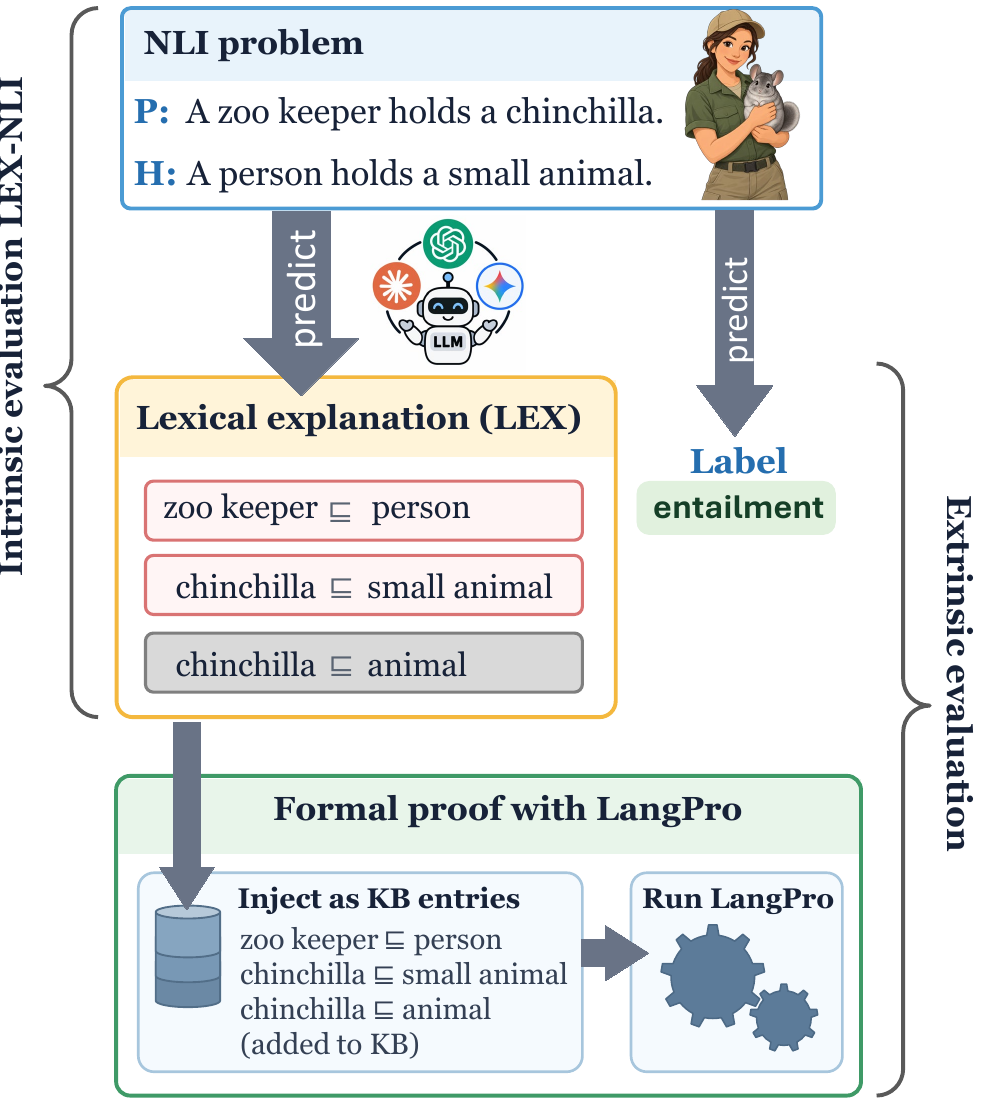}
    \caption{Overview of the proposed intrinsic and extrinsic evaluation. 
    Given an NLI problem, an LLM is jointly evaluated on the predicted inference label and a structured lexical explanation (\LEX{}) against human-curated multi-reference \LEX{} sets. 
    Extrinsically, the generated lexical relations are supplied to the LangPro theorem prover, which determines whether they suffice to construct a formal proof.}
    \label{fig:intro}
\end{figure}

To study how LLMs reason in NLI more closely, researchers have built simplified synthetic NLI-style datasets (that usually require multi-step reasoning) and pursued two related lines of work:
(i) train and evaluate LLMs on multi-step reasoning in natural language \cite{ruletakers:2020,zhong-etal-2022-analytical,saparov2023language,han-etal-2024-folio},
and 
(ii) use LLMs to encode sentence meaning in a formal logic formula that is later fed to rule-based reasoners, such as first-order logic theorem provers \cite{olausson-etal-2023-linc,pan-etal-2023-logic,ryu2025divide}. 
However, this suggests that the problems from the original NLI datasets%
\footnote{For instance, RTE \cite{Dagan:2006}, SICK \citep{marelli-etal-2014-sick}, and SNLI \citep{bowman-etal-2015-large}.
}
are too challenging for (i) and (ii), and such valuable datasets are unfortunately left out of these research lines.

To partially formalize NLI reasoning, we introduce an explainable task in
which models produce structured lexical relations supporting the inference
label. We also evaluate a neuro-symbolic pipeline distinct from (ii).
Specifically, LLMs generate lexical entailments---not logical formulas---to
help the natural logic prover LangPro \cite{abzianidze-2017-langpro} address
scarce KB resources.%
\footnote{
Unlike ours, the works in (i) and (ii) do not use LLMs to produce lexical
knowledge because their synthetic datasets require little covert lexical
knowledge.
}
This aligns with work treating LLMs as knowledge bases (KB)
\cite{petroni-etal-2019-language,alkhamissi2022reviewlanguagemodelsknowledge}.
Our evaluation setup is illustrated in \autoref{fig:intro}.

Our contributions are threefold:\\
\indent (1) We curate a test set of entailment problems together with the corresponding lexical entailments that are necessary and sufficient for solving them;\\
\indent (2) We evaluate multiple LLMs against the test set on the \LEX-NLI task, identifying relevant lexical entailments for NLI problems, and analyze the results with respect to model family and size;\\
\indent (3) We select high-performing LLMs and use them as KB generators for LangPro while the latter is evaluated on its ability to prove NLI problems. 
Together, these results demonstrate an interpretable integration of LLMs and a symbolic reasoner.

\section{Related Work}
\label{sec:rel-work}

NLI is one of the most widely studied NLP tasks.
The release of e-SNLI \citep{camburu-etal-2018} initiated a research line on explainable NLI by augmenting NLI problems with human-elicited free-text explanations and span highlights.
However, the lack of precise and adequate metrics for automatically comparing generated and reference explanations hinders progress in free-text explainable NLI.

Several alternatives to free-text explanations have therefore been explored.
The first line of work directly stems from e-SNLI and uses its span highlights for training and/or evaluation of explainable NLI models \cite{deyoung-etal-2020-eraser,zhao-vydiswaran-2021-lirex,carton-etal-2022-learn,stacey-etal-2022-logical,ghasemi-madani-minervini-2023-refer}.
However, the span highlights, which \newcite{camburu-etal-2018} used to facilitate free-text explanation collection, cannot be regarded as lexical knowledge and are too noisy for a gold standard.
The second line of work delegates reasoning to rule-based systems,
while LLMs translate sentences into meaning representations, often first-order logic \citep{olausson-etal-2023-linc,pan-etal-2023-logic,ryu2025divide,noble-etal-2025-mood}.
The third line of work constructs synthetic NLI-style datasets in which the reasoning process is represented as a tree, graph, or sequence of formally validated inference steps \citep{ruletakers:2020,zhong-etal-2022-analytical,saparov2023language,dalvi-etal-2021-explaining,han-etal-2024-folio}.
The last two lines of work rely on simpler language or more restricted reasoning patterns than those found in standard NLI benchmarks such as SNLI \citep{bowman-etal-2015-large} and SICK \citep{marelli-etal-2014-sick}.

In contrast, we use existing entailment problems from SNLI and SICK augmented with structured lexical explanations, formatted as curated multiple-reference relation sets.
We restrict the task to entailment rather than to synthetic sentences or simple reasoning patterns, as in the related work.

State-of-the-art rule-based NLI systems \citep{yanaka-etal-2018-acquisition,abzianidze-2020-learning} suffer from lexical knowledge sparsity despite using resources such as WordNet \citep{fellbaum1998wordnet} and ConceptNet \citep{speer2017conceptnet}.
Consequently, several studies used neural models to supply missing lexical/phrasal knowledge.
\citet{yoshikawa-etal-2019} applied the ComplEx KB completion model \citep{trouillon-etal-2016} to generate lexical relations for a theorem prover.
\citet{Tomihari-Yanaka-2023} used vision-language embeddings to assess phrase-level entailments selected by a logic-based NLI system.
For Japanese logic-based NLI, \citet{kobayashi-etal-2026-neural} developed a lexical-entailment classifier inspired by hyperbolic embeddings \citep{pmlr-v80-ganea18a}.
\citet{chen-etal-2021-neurallog} complemented WordNet and ConceptNet with phrasal paraphrases derived from a pretrained language model, while \citet{noble-etal-2025-mood} used an LLM to generate first-order-logic axioms corresponding to lexical relations for a theorem prover.
\citet{aly-etal-2023-qa} and \citet{strong-etal-2024-zero} on natural-logic-based fact verification use instruction-tuned LLMs to predict short-phrase relations for a version of natural logic.

Our work instead formulates lexical KB prediction as an independent explainable task.
Reference lexical relations are not tied to third-party systems or their representation language.
We judge them by whether they explain the NLI inference, not by whether a given reasoning system can import them and successfully use them in its proofs.

\section{Structured Explanation with Lexical Entailments}
\label{sec:lexerel}

We use lexical entailments, such as \entails{guitar}{musical instrument},
as structured NLI explanations.
We first motivate our focus on entailment problems and then show how lexical entailments can serve as a key component of the underlying reasoning.
Finally, we define the kinds of words and short phrases that may occur in lexical entailments.

Earlier formulations of Recognizing Textual Entailment \citep{Dagan:2006} distinguished between entailment and non-entailment, whereas contemporary NLI datasets typically divide non-entailment into contradiction and neutral.
The distinction between contradiction and neutral, however, is not always straightforward.%
\footnote{
This was already discussed by \citet{marelli-etal-2014-sick}.
Consider a premise-hypothesis pair from SICK: ``A couple is not looking at a map'' and ``A couple is looking at a map''.
The inference label depends on the interpretation of indefinite NPs: if ``a couple'' and ``a map'' co-refer across sentences, then the NLI problem is contradiction, otherwise neutral.
}
Moreover, non-entailment cases are generally more difficult to explain using only word- or phrase-level relations.
A neutral label usually results from the absence of certain semantic relations, and contradictions sometimes depend only on negation, which doesn't require relations.
In contrast, word- or phrase-level relations frequently represent knowledge needed to explain entailment NLI problems, e.g., ``a zoo keeper holds a chinchilla'' entailing ``a person holds a small animal'' is explained with \entails{zoo keeper}{person} and \entails{chinchilla}{small animal} entailment relations.
We therefore currently focus only on entailment problems and lexical entailment relations between words and multi-word lexical items.

A lexical entailment is a directed binary relation, e.g., \entails{guitar}{musical instrument}, where the first argument is semantically more specific than or equal to the second.
It is \emph{structured} because it represents the knowledge in an explicit format.
This makes the explanations easier to compare across annotators and models and allows them to be directly supplied to a symbolic reasoning system.%
\footnote{In contrast, a free-text explanation doesn't make the comparison straightforward: detecting equality between ``guitar means musical instrument'' and ``a guitar is a type of a musical instrument'' is not trivial.  
}

We use the term \emph{lexical} broadly enough to include both single- and multi-word concepts.
Lexical items are essentially similar to those found in lexical databases such as WordNet, but extending beyond their incomplete coverage.
Single-word relations include \entails{cap}{hat} and \entails{obese}{fat}, whereas multi-word relations include \entails{female swimmer}{woman} and \entails{little boy}{child}.
Multi-word expressions are included when using only one of their components would produce an invalid or insufficient relation.
For instance, \entails{swimmer}{woman} is false, whereas \entails{female swimmer}{woman} is acceptable.
Similarly, \entails{chinchilla}{animal} is insufficient for a hypothesis referring to a ``small animal'', as the more informative relation \entails{chinchilla}{small animal} is required.

For consistency and comparability, expressions in a relation are normalized to their lemmatized forms and exclude determiners, auxiliary verbs, the infinitival marker ``to'', and prepositional phrases.
We also exclude relations that are merely obtained by dropping modifiers:
\entails{tall man}{man} is uninformative because entailment follows from the compositionality by discarding \textit{tall}.

We call a set of lexical entailments that serves as a lexical basis of
explanations for an NLI problem a \LEX{} set, or \LEX{}. Some entailment
problems don’t require lexical entailments. When the hypothesis follows through
syntactic and compositional reasoning, the \LEX{} set is empty. For example,
“A zoo keeper holds a chinchilla” entailing “A chinchilla is held” doesn’t
require lexical relations but is licensed by passive alternation and
sub-categorization of “hold”, which omits the agent.

\LEX{} sets are not intended to constitute complete formal proofs of entailment.
Instead, they serve as structured proxies for the lexical knowledge that licenses an entailment.
When non-empty, they represent the lexical component that a complete formal or structured explanation must account for.

\section{Curating a Test Set}
\label{sec:test_set}

\begin{table*}[t]
\centering
\scalebox{.75}{
\setlength{\tabcolsep}{4pt}
\renewcommand{\arraystretch}{1}
\begin{tabular}{@{}p{1.22\textwidth}@{}}

\toprule

{\centering
\textbf{NLI problem with three LEX annotations and an optional 4th alternative LEX}
\par}
\\

\midrule


\parbox[c]{0.68\linewidth}{%
    \textbf{P:} A man in a bright yellow shirt juggles while riding a unicycle.\\
    \textbf{H:} A man performs a juggling act on a unicycle.
}
\hfill
\parbox[c]{0.29\linewidth}{%
    \begin{tabularx}{\linewidth}{@{}
        *{3}{>{\centering\arraybackslash}X}
    @{}}
        \lexempty & \lexempty & \lexempty
    \end{tabularx}
}
\\

\midrule


\textbf{P:} A young boy wearing safety swimming gear and goggles is in a pool.
\hspace{1em}
\textbf{H:} A kid uses eye protection while in the water.
\\[3pt]

\begin{tabularx}{\linewidth}{@{}
    *{3}{>{\centering\arraybackslash}X}
@{}}

\lexbox{
    \entails{goggles}{eye protection};\\
    \entails{pool}{water}; \quad
    \entails{young boy}{kid}
}
&
\lexbox{
    \entails{goggles}{eye protection};\\
    \entails{pool}{water}; \quad
    \entails{young boy}{kid}
}
&
\lexbox{
    \entails{goggles}{eye protection};\\
    \entails{pool}{water}; \quad
    \entails{young boy}{kid}
}

\end{tabularx}
\\

\midrule


\textbf{P:} A woman observes a rusted antique car.
\hspace{1em}
\textbf{H:} A woman looks at an old car.
\\[3pt]

\begin{tabularx}{\linewidth}{@{}
    *{4}{>{\centering\arraybackslash}X}
@{}}

\lexbox{
    \entails{antique}{old};\\
    \entails{observe}{look}
}
&
\lexbox{
    \entails{antique car}{old car};\\
    \entails{observe}{look}
}
&
\lexbox{
    \entails{antique}{old};\\
    \entails{observe}{look at}
}
&
\lexbox[dottedlex]{
    \entails{antique car}{old car};\\
    \entails{observe}{look at}
}

\end{tabularx}
\\

\midrule


\parbox[c]{0.68\linewidth}{%
    \textbf{P:} A person rides his bicycle in the sand beside the ocean.\\
    \textbf{H:} A person is on a beach.
}
\hfill
\parbox[c]{0.29\linewidth}{%
    \begin{tabularx}{\linewidth}{@{}
        *{3}{>{\centering\arraybackslash}X}
    @{}}
        \lexbox{\entails{sand}{beach}}
        &
        \lexempty
        &
        \lexempty
    \end{tabularx}
}
\\

\midrule

%
%

\parbox[c]{0.57\linewidth}{%
    \textbf{P:} A black and white dog is carrying a huge stick on the green grass.\\
    \textbf{H:} A black and white dog with a large branch is running in the field.
    \par\vspace{3pt}

    \begin{tabularx}{\linewidth}{@{}
        *{2}{>{\centering\arraybackslash}X}
    @{}}
        \lexbox{\entails{huge stick}{large branch}}
        &
        \lexbox[dottedlex]{\entails{huge}{large}}
    \end{tabularx}
}
\hfill
\parbox[c]{0.40\linewidth}{%
    \begin{tabularx}{\linewidth}{@{}
        *{2}{>{\centering\arraybackslash}X}
    @{}}

        \lexbox{
            \entails{green grass}{field};\\
            \entails{huge}{large}
        }
        &
        \lexbox{
            \entails{grass}{field};\\
            \entails{huge}{large};\\
            \entails{stick}{branch}
        }

    \end{tabularx}
}
\\

\bottomrule

\end{tabular}
}
\caption{Examples of agreement and variation among three independently produced \LEX{} annotations.
An additional 4th \LEX{}, if introduced during adjudication, is marked with a dashed borderline.
$\varnothing$ stands for an empty \LEX{}.
The first two problems have unanimous agreement in \LEX{} annotations. 
}
\label{tab:lex-annotation-examples}
\end{table*}

To evaluate models on explainable NLI from a lexical perspective, we curate a set of NLI problems paired with reference \LEX{} sets.
We constructed a manually annotated test set of entailment problems drawn from SICK and SNLI, as these datasets heavily depend on lexical knowledge.
All four authors served as both data curators and annotators.
Each annotator initially selected 70-100 entailment problems from the two datasets and assigned a \LEX{} set to each selected problem.
During this selection process, the annotators aimed for approximately 20-30\% of the problems to have an empty \LEX{}.
This ensured that the test set contained both problems requiring non-trivial lexical knowledge and problems whose entailment could be explained through syntactic and compositional reasoning alone.
All annotations were produced according to the guidelines presented in \autoref{app:annotation-prompt}.
The selected NLI problems were subsequently pooled into a single collection.
Each problem was then independently annotated by two additional annotators, besides its original annotator.
Consequently, every problem received three independently produced \LEX{} annotations.
This procedure allowed us to assess both the reproducibility of the task and the extent to which different annotators identify the same lexical knowledge as explanatory.

We considered the annotations to be in unanimous agreement when all three annotators assigned the same \LEX{} set.
During the first phase, out of the 365 annotated problems, 161 (44\%) received unanimous \LEX{} annotations.
The remaining cases exhibited variation in the selected relations, their granularity, or the amount of lexical knowledge considered necessary for explaining the inference.
These disagreements do not always indicate annotation errors, since more than one set of lexical entailments may provide an acceptable explanation for the same NLI problem (see \autoref{tab:lex-annotation-examples}).

We therefore conducted an adjudication phase to distinguish annotation mistakes from legitimate explanatory variation.
As a result, we obtained unanimous agreement for 223 (62\%) problems (see \autoref{tab:data-iaa}).%
\footnote{Three SNLI problems were excluded from the final dataset: one contained an incomplete sentence, and two had entailment labels that were judged unreliable.
}
When a disagreement could be resolved into a single adjudicated annotation, one annotator proposed the revised \LEX{}.
The revised annotation was subsequently checked by the annotators whose original \LEX{} sets had been modified.
This process ensured that changes were not introduced unilaterally and that the final annotation reflected agreement among the affected annotators.

\begin{table}[b!]
\centering
\small
\setlength{\tabcolsep}{3.5pt}
\scalebox{.9}{
\begin{tabular}{lrrrr}
\toprule
\multicolumn{5}{l}{\textit{Dataset statistics}} \\
\midrule
\multicolumn{4}{l}{Total NLI problems} & 362 \\
\multicolumn{4}{l}{SNLI problems / SICK problems} & 280 / 82 \\
\multicolumn{4}{l}{Problems w/ unique \LEX{} (unanimous agreement)\kern-5mm} & 223 (61.6\%) \\
\multicolumn{4}{l}{Problems with an alternative (4th) \LEX{}} & 22 (6.1\%) \\
\multicolumn{4}{l}{Total \LEX{} (incl. the added 4th \LEX{})} & 1108\\
\multicolumn{4}{l}{\LEX{} size distribution: 0, 1, 2, 3+} & 29 / 51 / 14 / 6\%\\
\midrule
\multicolumn{2}{l}{\textit{Annotator performance}} & $n$ & Exact Match\kern-5mm & \kern-8mmmicro-F1\\
\midrule
\multicolumn{2}{l}{Annotator-1 vs.\ rest} & 271 & 75.7 & 81.5 \\ %
\multicolumn{2}{l}{Annotator-2 vs.\ rest} & 271 & 84.5 & 86.0 \\ %
\multicolumn{2}{l}{Annotator-3 vs.\ rest}  & 271 & 80.8 & 85.9 \\ %
\multicolumn{2}{l}{Annotator-4 vs.\ rest} & 273 & 82.8 & 86.2 \\ %
\midrule
\multicolumn{2}{l}{\textit{Mean}} & -- & 80.9 & 84.9 \\
\bottomrule
\end{tabular}
}
\caption{Statistics of the final dataset and annotator performance after the adjudication. 
Annotator-N vs rest is based on the problems annotated by Annotator-N.
}
\label{tab:data-iaa}
\end{table}

When disagreement couldn't be reduced to a single canonical \LEX{} (in total 139), several \LEX{} variants were retained as a reference answer for the same problem.
In several instances, adjudication also resulted in the creation of a fourth \LEX{} variant that captured an acceptable explanation not represented by any of the three original annotations.
This outcome indicates that variation is, to some extent, inherent to the task rather than being solely a consequence of inconsistent annotation.
The examples from the final dataset are given in \autoref{tab:lex-annotation-examples}.

The resulting test set attaches one to four reference \LEX{} sets (possibly empty \textsc{lex}) to each NLI problem.
20\% of the \LEX{} annotations contain more than one relation, while 29\% are empty.
Overall annotator performance is 80.9\% for exact match and 84.9\% for micro-F1, computed as the mean of the four annotator-level scores.
For each annotator, we calculate the scores by comparing the annotator's \LEX{} to the other two or three \LEXs{} (depending on the presence of the 4th \LEX{}).%

\section{Lexical Entailments for Formal Proofs}
\label{sec:lex-to-proof}

The lexical entailments introduced in Section~\ref{sec:lexerel} can be converted directly into KB entries for logic-based NLI systems.
For example, \entails{chinchilla}{small animal} can be represented as a subsumption relation stating that every chinchilla is a small animal.
Such relations complement the logical rules of a prover with the lexical knowledge required to connect predicates occurring in the premise and hypothesis.
A \LEX{} set can therefore be viewed not only as a structured explanation of an inference, but also as a small problem-specific KB that may lead to a formal proof.

One logic-based NLI system that can use such knowledge is LangPro, a natural language theorem prover based on natural logic and semantic tableaux \citep{abzianidze-2017-langpro,abzianidzethesis}.%
\footnote{
LangPro analyzes the linguistic structures of the premise and hypothesis and converts them into logical forms that remain relatively close to the original sentences.
To test whether the premise entails the hypothesis, it attempts to construct a tableau from the premise together with the negation of the hypothesis.
The prover repeatedly applies logical, lexical, and monotonicity-based inference rules, and the entailment is established when every branch of the resulting tableau is closed by a contradiction.
The resulting tableau constitutes a formal proof whose individual inference steps can be inspected.
}
In addition to its logical rules, LangPro relies on a lexical KB containing relations such as synonymy, hypernymy, and semantic incompatibility.
A large part of this knowledge is obtained from WordNet.
However, WordNet does not contain every relation required by NLI problems.
Consequently, a tableau may remain open even when LangPro's linguistic analysis and proof rules are otherwise sufficient to derive the intended inference.
For instance, 
our running NLI example requires relations corresponding to \entails{zoo keeper}{person} and \entails{chinchilla}{small animal}, neither of which is available in WordNet.%
\footnote{Feeding LangPro with lexical relations is straightforward because multi-word expressions such as \entarg{zoo keeper} and \entarg{small animal} are already terms in its higher-order logic, without need to represent them as, for example, $\entarg{small}(x)\wedge\entarg{animal}(x)$.
}

We use LLMs to supplement LangPro with lexical knowledge that is missing from WordNet.
Rather than asking an LLM to predict the NLI label, construct the proof itself, or translate sentences into logic formulas, we use it only to propose relevant lexical knowledge.
After receiving lexical KB from an LLM, LangPro remains solely responsible for determining whether the supplied KB is sufficient to find a proof.
Thus, LangPro provides an extrinsic test of whether LLMs generate lexical entailments relevant for formal proof construction.

\section{Experiments}
\label{sec:experiments}

We conduct two groups of experiments:
evaluating LLMs intrinsically on the new task of predicting \LEX{} sets and extrinsically on how much those predictions help LangPro prove NLI problems.%
\footnote{The dataset and codebase are available from \url{https://github.com/jorrytdejong/kbprojection}.}

\subsection{Predicting Lexical Explanations}
\label{ssec:pred-lex}

We evaluate eight instruction-following LLMs from four model families:
GPT-5.4 and GPT-5.4 Mini, Claude Sonnet 4.5 and Claude Haiku 4.5, Gemini 3.5
Flash and Gemini 3.1 Flash-Lite, GPT-OSS-20B, and Gemma 3 4B. The selection
combines higher-capability hosted models with smaller, cheaper alternatives from the same families, allowing us to examine whether
\LEX{} prediction requires a frontier model or can be handled by cheaper models.
It also reduces dependence on a single developer.
Additionally, GPT-OSS-20B and Gemma 3 4B provide open-weight comparison points.

Each model is evaluated five times using the same prompt.
Only relations extracted from raw model responses are evaluated: 
no automatic post-processing was used to correct word forms or discard relations based on the vocabulary of an NLI problem.
LLM inference details are provided in \autoref{app:llm-inf-setup}.
The prompts follow the annotation guidelines (see \autoref{app:annotation-prompt}). %
Prompt selection details are in \autoref{app:prompt-selection}.

Predictions are evaluated against all reference \LEX{} sets, and the best match determines the final score. 
We use two scores to evaluate models on labeling entailment problems and generating \LEX{} sets.
\emph{Exact Match} (EM) counts a predicted \LEX{} set correct when it is equal to one of the reference \LEX{} sets. 
Because EM ignores the size of \LEX{}, we additionally use \emph{micro-F1} for finer-grained matching.%
\footnote{
We opt for micro-F1 because it gives greater weight to larger \LEX{} sets and is less influenced by the 29\% of examples with empty \LEX{} sets than macro-F1.
}

Mismatch between lexical entailment relations can be roughly divided into normalization, structure, and semantic errors.
\emph{Normalization errors} concern wrong lemmatization of words in the relations.
\emph{Structure errors} include cases where two correctly extracted arguments are placed in the wrong order in a relation, i.e., in a swapped order for the entailment relation. 
\emph{Semantic errors} cover mistakes that do not belong to the two aforementioned classes. 
To assess how much models struggle with normalization, we use \emph{lemmatized F1} that eliminates this type of error by lemmatizing \LEX{} sets before matching.%
\footnote{
We POS-tag the original premise and hypothesis and locate words of each argument in either sentence.
The corresponding tags are mapped to WordNet's noun, verb, adjective, or adverb categories,
and NLTK's WordNet lemmatizer is used to get the lemma for a word-tag pair. 
Tokens for which this procedure fails remain unchanged.
We also apply lemmatization to the reference relations because they contain plural-only nouns such as ``glasses'', ``pants'', and ``shades'', and this might increase the chance of a match. 
}
We also use \emph{argument-unordered F1}, which ignores relation direction during matching, so that \entails{guy}{man} matches \entails{man}{guy}.

\begin{table*}[t!]
\centering
\scalebox{.8}{
\begin{tabular}{lcccccccc}
\toprule
\multirow{2}{*}{\textbf{Model}} &
\multirow{2}{*}{\textbf{Support}} &
\multirow{2}{*}{\shortstack{\textbf{Exact}\\\textbf{match}}} &
\multicolumn{4}{c}{\textbf{Micro-F1}} &
\multirow{2}{*}{\shortstack{\textbf{Micro-}\\\textbf{Precision}}} &
\multirow{2}{*}{\shortstack{\textbf{Micro-}\\\textbf{Recall}}} \\
\cmidrule(lr){4-7}
& & & \textbf{Raw $\downarrow$}
& \textbf{$\Delta$ Lem.}
& \textbf{$\Delta$ Un.}
& \textbf{$\Delta$ Lem.+Un.}
& & \\
\midrule
Claude Sonnet 4.5
& \llap{1}00.0
& \textbf{58.2} {\small ($\pm$ 0.2)}
& \textbf{73.1} {\small ($\pm$ 0.2)}
& +0.9 & +0.6 & +1.5
& \textbf{64.5} {\small ($\pm$ 0.2)}
& \textbf{84.2} {\small ($\pm$ 0.3)} \\
Gemini 3.1 Flash-Lite
& \llap{1}00.0
& 57.9 {\small ($\pm$ 0.4)}
& 72.3 {\small ($\pm$ 0.5)}
& +0.4 & +0.0 & +0.4
& 64.4 {\small ($\pm$ 0.4)}
& 82.6 {\small ($\pm$ 0.6)} \\
Gemini 3.5 Flash
& \llap{1}00.0
& 58.1 {\small ($\pm$ 0.4)}
& 71.9 {\small ($\pm$ 0.4)}
& +1.0 & +0.4 & +1.4
& 63.6 {\small ($\pm$ 0.4)}
& 82.6 {\small ($\pm$ 0.5)} \\
\baseoracle{} WordNet \& Oracle
& \llap{1}00.0
& 64.1 \phantom{{\small ($\pm$ 0.0)}}
& 70.5 \phantom{{\small ($\pm$ 0.0)}}
& -- & -- & --
& \llap{1}00.0 \phantom{{\small ($\pm$ 0.0)}}
& 54.4 \phantom{{\small ($\pm$ 0.0)}}\\
GPT-5.4
& \llap{1}00.0
& 57.5 {\small ($\pm$ 1.0)}
& 69.3 {\small ($\pm$ 0.4)}
& +0.7 & +0.4 & +1.1
& 61.0 {\small ($\pm$ 0.4)}
& 80.2 {\small ($\pm$ 0.7)} \\
Claude Haiku 4.5
& \llap{1}00.0
& 50.6 {\small ($\pm$ 0.0)}
& 67.2 {\small ($\pm$ 0.0)}
& +1.6 & +0.0 & +1.6
& 57.1 {\small ($\pm$ 0.0)}
& 81.8 {\small ($\pm$ 0.0)} \\
GPT-5.4 Mini
& \llap{1}00.0
& 38.7 {\small ($\pm$ 1.2)}
& 56.0 {\small ($\pm$ 1.1)}
& +7.3 & +0.3 & +7.8
& 46.3 {\small ($\pm$ 0.9)}
& 71.0 {\small ($\pm$ 1.5)} \\
GPT-OSS-20B
& 99.8
& 44.0 {\small ($\pm$ 1.1)}
& 55.0 {\small ($\pm$ 0.9)}
& +4.5 & +0.2 & +4.7
& 48.8 {\small ($\pm$ 0.7)}
& 63.0 {\small ($\pm$ 1.3)} \\
\baseoracle{} WordNet \& Rules
& \llap{1}00.0
& 50.0 \phantom{{\small ($\pm$ 0.0)}}
& 51.0 \phantom{{\small ($\pm$ 0.0)}}
& -- & -- & --
& 70.2 \phantom{{\small ($\pm$ 0.0)}}
& 40.0 \phantom{{\small ($\pm$ 0.0)}}\\
Gemma 3 4B
& 98.1
& 14.1 {\small ($\pm$ 0.3)}
& 30.6 {\small ($\pm$ 0.3)}
& +11.5 & +0.1 & +12.1
& 21.4 {\small ($\pm$ 0.2)}
& 53.8 {\small ($\pm$ 0.3)} \\
\baseoracle{} Empty LEX
& \llap{1}00.0
& 37.0 \phantom{{\small ($\pm$ 0.0)}}
& \phantom{0}0.0 \phantom{{\small ($\pm$ 0.0)}}
& -- & -- & --
& \llap{1}00.0 \phantom{{\small ($\pm$ 0.0)}}
& \phantom{0}0.0 \phantom{{\small ($\pm$ 0.0)}}\\
\bottomrule
\end{tabular}
}
\caption{
LLMs' results on the \LEX{}-NLI task.
Values are reported as the mean {\small ($\pm$ sample standard deviation)} across five runs.
Results are sorted by raw Micro-F1, computed without post-processing.
$\Delta$ Lem., $\Delta$ Un., and $\Delta$ Lem.+Un. report gains over raw Micro-F1 under lemmatized, argument-order-agnostic (\entails{A}{B} matches \entails{B}{A}), and combined scoring, respectively.
Support is the percentage of problems evaluated; 
exclusions are due to error responses from API calls.
\baseoracle{} denotes baseline and oracle systems.
}
\label{tab:lex_prediction}
\end{table*}

\subsection{Supplying Knowledge to LangPro}
\label{ssec:llm-langpro}

For the extrinsic evaluation, we use an LLM to supply lexical entailments as a KB to LangPro while it searches for proofs of entailment problems. 
1,000 randomly sampled entailment problems from the SNLI training set are used for this evaluation.
We compare five ways of assisting LangPro with lexical KBs: only WordNet, only LLM-generated knowledge, and WordNet with LLM-generated knowledge, with the latter two settings evaluated both in one-shot and agentic use of LLMs.

In the one-shot setting, the LLM generates a KB once and LangPro attempts a proof using the KB.
In the agentic setting, a failed proof triggers one refinement cycle.
A critic prompt receives the premise-hypothesis pair together with LangPro's proof outcome, tableau closure information, and the attempted KB, and diagnoses possible missing or inappropriate relations.
A subsequent LLM call uses this diagnosis to generate a revised KB, after which LangPro attempts the proof again. 
The experiment therefore tests whether feedback from the symbolic prover, followed by a critical reassessment of the generated KB, can improve the usefulness of LLM-generated lexical knowledge on a second attempt.

\section{Results}
\label{sec:results}

\subsection{Predicting Lexical Explanations}
\label{ssec:results-lex}

The evaluation results of the eight models on the curated 362 problems are given in \autoref{tab:lex_prediction}.
The evaluation is averaged over five runs, and the scores (with micro-F1 regarded as primary) are computed based on the best reference \LEX{} sets.
Since the proposed explainable NLI task requires models to predict an inference label together with a \LEX{} set, any \LEX{} set paired with a non-entailment prediction is considered distinct from a gold empty \LEX{}.
However, if a model's prediction for a problem is not available due to an error in the API call, we ignore such problems from the model evaluation (indicated in the support column).

In addition to the eight LLMs, we report results for three baselines.
Empty \LEX{} always returns an empty \LEX{} set and serves as the simplest baseline.
WordNet\,\&\,Rules returns relations of the form \entailsnormal{$s_p$}{$s_h$} that are found in WordNet, where $s_p$ and $s_h$ are single- or multi-word lemma spans that occur only in the premise and hypothesis, respectively.%
\footnote{
WordNet\,\&\,Rules represents a realistic rule-based baseline using the heuristic that the first and second components of a relation typically come from the premise and hypothesis, respectively, while words shared by the premise and hypothesis are rarely part of such relations.
It uses spaCy \citep{honnibal2020spacy} for lemmatization and POS tagging, with POS tags used to select the corresponding WordNet senses.
During WordNet lookup, we use the transitive closure of hypernymy, entailment, and similarity relations.
}
WordNet\,\&\,Oracle peeks at the reference \LEX{} sets and selects all and only those relations that are validated by WordNet lookup.
It therefore represents a baseline that is perfect in selection of arguments, but limited by WordNet coverage.

Model performance ranking generally follows a simple size-based pattern, except for Gemini 3.1 Flash-Lite
which achieves better scores than Gemini 3.5 Flash. 
Interestingly, Gemini 3.1 Flash-Lite also substantially outperforms Claude Haiku 4.5 and GPT-5.4 Mini, from a comparable product tier. 
GPT-5.4 Mini, GPT-OSS-20B, and Gemma 3 4B scored substantially lower than the remaining models, with Gemma 3 4B performing worst overall and falling below the WordNet\,\&\,Rules baseline on both EM and micro-F1, as well as below the Empty LEX baseline on EM.

The substantially higher recall than precision suggests systematic overgeneration.
Across the top five models, predicted \LEX{} sets average 1.4 relations, compared with 1.0 for the reference \LEX{} sets.
Relative to the references, models produce fewer empty sets (15\% vs 29\%) and more sets of size 2 (25\% vs 14\%) and 3+ (11\% vs 6\%).
Due to the models' tendency to overgenerate relations, WordNet\,\&\,Oracle provides a strong baseline.
This suggests that selecting the correct lexical spans is a major challenge of the task.

\autoref{tab:lex_prediction} also reports micro-F1 under lemmatized and/or argument-order-agnostic matching.
Ignoring inflectional differences increases micro-F1 for all models, but the gain is small for the top five models ($\mathord{\leq}1.6$ points) and substantially larger for the remaining models ($\mathord{\geq}4.5$ points).
This suggests that producing lemmatized arguments is not a major difficulty for the better-performing models.

Ignoring argument order has a much smaller effect ($\mathord{\leq}0.6$ points).
We hypothesize that this reflects a combination of models distinguishing more specific from more general concepts and exploiting the heuristic that the specific concepts usually occur in the premise and the general ones in the hypothesis.
Note that among the eight entailment examples in the nine-shot prompt, we deliberately included one example that violates this heuristic (Example~7 in \autoref{app:annotation-prompt}).
However, because entailment NLI problems rarely include such adversarial cases, we included only one example in the prompt.

\begin{table*}[t]
\centering
\scalebox{.9}{
\begin{tabular}{l
    c@{\hspace{2em}}
    c@{\hspace{2em}}
    c@{\hspace{2em}}
    c}
\toprule

\multirow{2}{*}{\textbf{Model}} &
\LLMone &
\textbf{\LLMa} &
\textbf{\LLMWN} &
\textbf{\LLMaWN} \\

&
\dWNhead
&
\dWNhead
\dgap
\dAhead
&
\dWNhead
&
\dWNhead
\dgap
\dAhead
\\

\midrule

Gemini 3.1 Flash-Lite
& \textbf{165}\dgap\dWN{+8}
& 180\dgap\dWN{+23}\dgap\dA{+15}
& 179\dgap\dWN{+22}
& 193\dgap\dWN{+36}\dgap\dA{+14} \\

Claude Haiku 4.5
& \textbf{165}\dgap\dWN{+8}
& 176\dgap\dWN{+19}\dgap\dA{+11}
& 179\dgap\dWN{+22}
& 190\dgap\dWN{+33}\dgap\dA{+11} \\

GPT-5.4 Mini
& 162\dgap\dWN{+5}
& 183\dgap\dWN{+26}\dgap\dA{+21}
& 178\dgap\dWN{+21}
& 199\dgap\dWN{+42}\dgap\dA{+21} \\

GPT-OSS-20B
& 158\dgap\dWN{+1}
& \textbf{194}\dgap\dWN{+37}\dgap\dA{+36}
& 174\dgap\dWN{+17}
& \textbf{212}\dgap\dWN{+55}\dgap\dA{+38} \\

Gemma 3 4B
& 163\dgap\dWN{+6}
& 163\dgap\dWN{\phantom{0}+6}\dgap\dA{\phantom{0}+0}
& \textbf{182}\dgap\dWN{+25}
& 182\dgap\dWN{+25}\dgap\dA{\phantom{0}+0} \\

\bottomrule
\end{tabular}
}
\caption{
LangPro proof coverage on 1,000 SNLI entailment problems when using various sources of the lexical KB.
\LLMone{} and \LLMa{} denote one-shot and agentic LLM-generated LEX, while \LLMWN{} and \LLMaWN{} are the former two coupled with WordNet, respectively.
Values indicate the number of problems proved as entailment.
The \underline{LangPro+WordNet baseline proves 157 problems}.
The boxed $\Delta_{\text{\textsc{wn}}}$ values indicate gains over LangPro+WordNet, while $\Delta_a$ values indicate the additional gains from agentic inference over the corresponding non-agentic LLM.
}
\label{tab:langpro-1000-results}
\end{table*}

Overall, the results show substantial room for improvement on the \LEX{}-NLI task.
The best model trails annotator performance by 22.7 points in EM and 11.8 points in micro-F1.
Even under the more model-friendly evaluation, where problems classified as non-entailment are excluded rather than penalized (see \autoref{tab:lex_prediction_friendly} in \autoref{app:mix}), the best scores remain below average annotator performance, with gaps of 13.9 percentage points in EM and 7.1 in micro-F1.%
\footnote{
Additionally, while prompt engineering was not extensive, the selection was biased toward the test set; see \autoref{app:prompt-selection}.
}

\subsection{Using \LEX{} to Prove Entailments}
\label{ssec:proof-results}

Table~\ref{tab:langpro-1000-results} reports the number of entailment proofs constructed by LangPro for 1,000 SNLI entailment problems.
When using only WordNet as its lexical KB, LangPro proves 157 problems.
We additionally evaluate four configurations in which the lexical KB is generated by an LLM either in one-shot mode (\LLMone{}) or agentic mode (\LLMa{}), with the generated KB optionally complemented by WordNet (\LLMWN{} and \LLMaWN{}, respectively).
Due to budget constraints, we conduct this experiment with five small models.

The extrinsic evaluation results show that the model ranking differs from that observed on the \LEX{}-NLI task.
While Gemini 3.1 Flash-Lite and Claude Haiku 4.5 remain the strongest models in the \LLMone{} setting, they are outperformed by other models in the remaining three configurations.

The relatively small gains of \LLMone{} over the LangPro+WordNet baseline (see $\Delta_{\text{\textsc{wn}}}$ values in \LLMone{}) do not appear to arise simply because the LLM-generated relations are largely subsumed by WordNet.
When combined with WordNet, each model still generates relations enabling additional proofs, see the $\Delta_{\text{\textsc{wn}}}$ values in the \LLMWN{} and \LLMaWN{} columns.
In the one-shot setting with WordNet (\LLMWN{}), Gemma 3 4B performs best, with 182 proofs, whereas GPT-OSS-20B performs best in the agentic setting \LLMaWN{}, with 212 proofs.

Agentic generation improves proof coverage for all models except Gemma 3 4B, indicating that Gemma cannot successfully revise the relations produced in its initial attempt.
GPT-OSS-20B benefits most from agentic generation, improving over the corresponding one-shot configurations by 36 proofs without WordNet and 38 proofs with WordNet.
It consequently achieves the best overall result, with 212 proofs in the \LLMaWN{} configuration.

We also conducted an extrinsic evaluation on the \LEX{} test set (see \autoref{tab:langpro-362-results} in \autoref{app:mix}).
The performance gaps between models are substantially narrower than on \LEX{}-NLI, indicating that the extrinsic evaluation is coarser-grained and captures a different aspect of model quality.
The \LEX{} test set also appears easier: proof coverage is typically around 1.5-1.9 times as high as on the 1,000-problem SNLI sample, while gains over the WordNet baseline are substantially larger.
In the \LLMaWN{} setting, all but one model exceed the 34\% accuracy of the LangPro-\LEX{} oracle, where LangPro is supplied with the union of the reference \LEX{} sets.
This suggests that, despite suboptimal intrinsic performance on \LEX{}-NLI, LLM-generated lexical knowledge can have an extrinsic effect comparable to the curated reference \LEX{} sets.

\section{Analysis}
\label{sec:analysis}

\subsection{Predicting Lexical Explanations}
\label{ssec:analysis-lex}

We examine the NLI problems and the predicted \LEX{} sets to understand the type of mistakes models make.
For 33 problems (9\%), no model predicts any of the reference \LEX{} sets exactly.
Our analysis of these cases yields three main observations.

\textbf{Overgeneration of variants:}
Models frequently overgenerate variants of the same semantic relation (in roughly half of the cases), most often differing only in inflection.
Examples include \entails{shades}{glasses} alongside \entails{shade}{glass}, \entails{cartoon}{animated} alongside \entails{cartoon}{animate}, and \entails{stereo equipment}{audio equipment} alongside \entails{stereo}{audio}.
Such overgeneration lowers precision and micro-F1 but does not affect recall.
The extent of this penalty is reflected in the improvement obtained with lemmatized micro-F1 (\autoref{tab:lex_prediction}).
Additional prompt instructions could likely reduce variant overgeneration, although this would not improve recall.

\textbf{Overgeneration of phrasal relations:}
\LEX-NLI specifically targets lexical relations and excludes phrasal ones. For example, for the entailment snippets ``man painting a picture'' and ``man is creating art'', it expects the explanation $\{\entails{paint}{create}, \entails{picture}{art}\}$.
However, no model predicted either relation, often opting instead for phrasal relations such as \entails{paint}{create art} or \entails{paint picture}{create art}, more natural in this case.  
Although the lexical-phrasal distinction is relatively clear from a syntactic perspective, the annotation guidelines, which were also used as the prompt, deliberately avoided syntax-heavy terminology.

\textbf{Issues with references:}
Several correct predictions were unfairly penalized because the references contained wrong relations instead of correct ones (\entails{look out}{view} instead of \entails{look out over}{view}), omitted valid ones (\entails{make}{cause} as in ``make laugh'' and ``cause to be merry''), or lacked acceptable variants (having \entails{decorated}{nice}, but not \entails{highly decorated}{nice}).

\subsection{Supplying Knowledge to LangPro}
\label{ssec:analysis-proof}

To assess the additional contribution of the agentic mode over one-shot generation, we manually analyzed the 15/14 additional problems solved by \LLMa{}/\LLMaWN{} compared with \LLMone{}/\LLMWN{} for Gemini 3.1 Flash-Lite.
The added relations for these problems are shown in \autoref{tab:gemini-agentic-quality-plus15} and \autoref{tab:gemini-agentic-quality-plus14}, respectively, in \autoref{app:mix}.
Only five of the analyzed problems contain at least one relation we judged semantically reasonable in isolation.
In the remaining cases, the additional relations help LangPro prove the problem either by compensating for errors introduced by the CCG parser or by exploiting weaknesses in LangPro's representation or proof procedure.
For example, for the problem P: ``Two skiers stand near a snowy mountain.'' and H: ``Some humans standing.'', LangPro fails to construct a proof even when \LLMWN{} provides \entails{skier}{human}.
This is because ``standing'' in the hypothesis is analyzed as a noun with lemma ``standing'' by the NLP tools used in LangPro's syntactic pipeline.
When \LLMaWN{} additionally proposes \entails{stand}{standing} and \entails{standing}{stand}, LangPro succeeds; the resulting proof requires only \entails{stand}{standing} together with \entails{skier}{human}.
Another type of tool error that these relations can compensate for involves confusing particles and prepositions.
Such cases illustrate that the agentic mode can improve proof success by repairing parser-induced lexical mismatches when supplying reasonable lexical knowledge is not enough.

\section{Conclusion \& Future Work}

To our knowledge, this is the first work that augments existing naturalistic NLI problems with structured lexical explanations of the knowledge required for their reasoning.
Unlike prior lexical relation detection work \citep{vulic-etal-2017-hyperlex,schmitt-schutze-2019-sherliic,shwartz-dagan-2016-adding}, among others, our task requires identifying only the relations needed to explain a particular inference.
We operationalize these explanations as \LEX{} sets and construct a manually annotated test set that retains multiple references when necessary.

Our intrinsic evaluation on \LEX-NLI shows that LLMs can recover much of the required lexical knowledge, but the task remains challenging.
Under the primary evaluation, the best-performing LLM reaches 73.1 micro-F1 and 58.2 EM (\autoref{tab:lex_prediction}), trailing average annotator performance by 11.8 and 22.7 points, respectively. 
Under the more model-friendly evaluation, which excludes problems predicted as non-entailment, the best per-metric scores increase to 77.8 micro-F1 and 67.0 EM (\autoref{tab:lex_prediction_friendly}), reducing the corresponding gaps to 7.1 and 13.9 points. 
Models particularly struggle with overgenerating relations, including inflectional variants, and with selecting minimal argument spans that constitute valid lexical entities.

Our extrinsic evaluation shows that LLM-generated KBs can enable LangPro, a natural-logic theorem prover, to find additional formal proofs.
LLMs can supply lexical relations that are absent from WordNet, while the agentic setting, which revises an initial KB once, further improves proof coverage.
However, much of this additional gain comes from semantically unsound relations that compensate for errors in the syntactic analysis of the input.
On the \LEX{} test set, the agentic setting even surpasses LangPro supplied with oracle reference relations, illustrating that higher proof coverage does not necessarily imply better lexical knowledge.

Overall, the results support a neuro-symbolic division of labor in which LLMs supply flexible lexical knowledge and symbolic systems perform explicit reasoning, while structured lexical evaluation provides an independent check on whether the supplied knowledge genuinely explains the inference.

As future work, we plan to scale up the \LEX{}-NLI task to cover all SICK problems and a larger portion of SNLI.
To make this feasible, we will combine manual validation with automatic candidate-\LEX{} detection using LangPro supplied with LLM-generated lexical knowledge.
As the analysis of model outputs in \autoref{ssec:analysis-lex} reveals, a few overgenerated relations are in fact valid; we therefore plan to use them to refine the reference annotations and improve their completeness as we develop the dataset.
In the longer term, we aim to extend structured NLI explanations beyond entailment to other NLI labels and to richer explanatory structures that capture reasoning beyond lexical entailment as outlined in \citet{abzianidze2025formalproofsstructuredexplanations}.

\newpage
\section*{Limitations}

Our study is restricted to NLI problems with an entailment label.
This choice allows us to study lexical entailments in a relatively well-defined setting, but leaves contradiction and neutral cases unexplored.
Extending structured explanations to these labels is less straightforward, as their reasoning may depend on incompatibility, negation, missing information, or pragmatic assumptions rather than directed lexical entailments alone.

We also deliberately restrict explanations to lexical entailments over words and multi-word concepts.
Such relations often constitute an important component of entailment reasoning, but they do not provide a complete account of the inference.
\LEX{} sets abstract away from syntactic reasoning, argument structure, quantification, coreference, event relations, and other compositional or world-knowledge phenomena.
Our results should therefore be interpreted as evaluating one crucial structured component of explanation rather than complete explanations of NLI reasoning.

Finally, the annotation task itself also admits genuine explanatory variation, as evidenced by cases in which several \LEX{} variants remained valid after adjudication.
Although our multi-reference evaluation (with the 4th \LEX{} added during the annotation adjudication phase) accounts for the retained alternatives, the reference set may not exhaust all semantically acceptable explanations.

\section*{Acknowledgments}

We thank the three anonymous reviewers for their constructive feedback and helpful suggestions.
This work was supported by the Dutch Research Council (NWO) through the project \emph{Towards Explainable Natural Language Reasoning for Artificial Intelligence} (project number \href{https://doi.org/10.61686/DXQIL95775}{DXQIL95775}).
This paper grew out of a student project conducted within the course \emph{Reasoning with Natural Language} (TLMV25020) at Utrecht University.

\bibliography{custom}

\appendix
\clearpage

\section{LLM Inference Setup}
\label{app:llm-inf-setup}

The models were accessed through the OpenRouter Chat Completions interface.
The model identifiers record the versions requested from OpenRouter, but they are not pinned provider snapshots.
The underlying snapshots were not retained in the response metadata.
In the final model comparison, every model was evaluated on the test data five times using the same nine-example few-shot prompt (see \autoref{app:annotation-prompt}).
Repetition was used to quantify variation caused by nondeterministic model and provider behaviour.
A temperature of 0 was requested, but this did not guarantee identical outputs because temperature support was not uniform across the evaluated OpenRouter routes and the setting could be ignored by a provider. Requests were issued with a concurrency of four, a 120-second timeout, and at most two retries. 
No common \texttt{top\_p}, random seed, reasoning-effort setting, stop sequence, or explicit output-token limit was supplied.
The total budget of the experiments was under \$80.

\section{Annotation Guidelines and LLM Prompt}
\label{app:annotation-prompt}

The following instructions were used both as annotation guidelines for constructing the gold-standard structured explanations and as the prompt given to the evaluated LLMs.
The placeholders \texttt{\$PREMISE} and \texttt{\$HYPOTHESIS} were replaced with the premise and hypothesis of each NLI problem.

Following \autoref{sec:lexerel}, the guidelines define a \LEX{} as the smallest set of directed lexical entailment relations
needed to explain how the hypothesis follows from the premise, when it does. 
The instructions require short, factually acceptable argument phrases and
emphasize four properties: 
(a) \emph{minimality} requires the model to omit relations unnecessary for the inference or that follow merely by discarding a modifier.
(b) \emph{lemmatization} requires canonical word forms without determiners, auxiliary verbs, or the infinitival marker \emph{to}. 
(c) \emph{relation direction} requires the first argument to license the second in the direction needed for the inference. 
(d) \emph{output formatting} requires the prescribed format for non-empty relation sets.

\begin{promptbox}[Lexical-entailment annotation prompt]

You are an expert in linguistic semantics and logic.
You will receive a Natural Language Inference (NLI) problem in English, consisting of a premise sentence and a hypothesis sentence.

You will reason carefully and decide whether the premise entails the hypothesis.
Entailment means that if the premise is true, then the hypothesis must also be true under ordinary English meaning and widely accepted background knowledge.

If the answer is \texttt{entailment}, output a structured explanation consisting of a set of lexical entailment relations over short phrases that explain why the hypothesis is entailed by the premise.
A lexical entailment must be defined over short phrases that are lemmatized or normalized versions of expressions occurring in the premise and hypothesis.
Use the format \entails{phrase\_1}{phrase\_2}.
The relation means that \textit{phrase\_1} is a type of, or semantically entails, \textit{phrase\_2}.
Examples include \entails{woman}{person}, \entails{dog}{domestic animal}, \entails{huge}{very big}, and \entails{run}{move fast}.

Use lexical entailment relations only when they are needed to explain the entailment.
If the entailment follows without any non-trivial lexical relation, output an empty set.

\medskip
\textbf{Relation-formatting rules}

\begin{enumerate}[leftmargin=5mm,itemsep=2pt,topsep=3pt]
    \item The meaning of a lexical entailment relation must be acceptable according to ordinary common-sense knowledge.
    For example, \entails{woman}{blond person} is not acceptable.

    \item A lexical entailment may not express a trivial relation that can be obtained simply by discarding modifiers.
    For example, \entails{blond woman}{woman} is not acceptable.

    \item A lexical entailment may not contain redundant words such as auxiliary verbs or the infinitival marker \textit{to}.
    For example, \entails{will walk}{will move}, \entails{is red}{is colored}, and \entails{to walk}{to move} are not acceptable.

    \item The words in each phrase must be lemmatized.
    For example, \entails{dogs}{domestic animals} is not acceptable.

    \item Phrases may not contain determiners.
    For example, \entails{a dog}{a domestic animal} is not acceptable.

    \item Phrases may not contain prepositional phrases.
    For example, \entails{dog with spots}{domestic animal} is not acceptable.
\end{enumerate}

\medskip
\textbf{Example 1}

\promptinput
    {Young ladies are playing the guitar.}
    {A musical instrument is being played by girls.}

\textbf{Correct output:}

\promptanswer{entailment}
\promptrelations{\{\entails{young lady}{girl}, \entails{guitar}{musical instrument}\}}

\textbf{Unwanted output:}

\promptrelations{\{\entails{young ladies}{girls}, \entails{the guitar}{a musical instrument}\}}

\textbf{Explanation:}
Phrases in relations may not contain determiners such as \textit{a} and \textit{the}.
The words \textit{ladies} and \textit{girls} must be replaced by their lemmas, \textit{lady} and \textit{girl}, respectively.

\medskip
\textbf{Example 2}

\promptinput
    {A female swimmer getting out of the pool still dripping wet.}
    {A woman gets out of the pool.}

\textbf{Correct output:}

\promptanswer{entailment}
\promptrelations{$\{\entails{female swimmer}{woman}\}$}

\textbf{Unwanted output:}

\promptrelations{$\{\entails{swimmer}{woman}\}$}

\textbf{Explanation:}
The proposed relation is factually incorrect because not every swimmer is a woman.

\medskip
\textbf{Example 3}

\promptinput
    {A young girl wearing a pink coat plays with a yellow toy.}
    {A kid is swinging a toy golf club.}

\textbf{Correct output:}

\promptanswer{non-entailment}

\medskip
\textbf{Example 4}

\promptinput
    {A black race car starts up in front of a crowd of people.}
    {A car is running.}

\textbf{Correct output:}

\promptanswer{entailment}
\promptrelations{$\{\entails{start up}{run}\}$}

\textbf{Unwanted output:}

\promptrelations{\{\entails{starts up}{is running}\}}

\textbf{Explanation:}
The expressions \textit{starts up} and \textit{is running} are not lemmatized.
The auxiliary verb \textit{is} is also unnecessary.

\medskip
\textbf{Example 5}

\promptinput
    {A woman dressed in red clothing is dancing inside a crowd of people.}
    {A woman in red is dancing in a crowd.}

\textbf{Correct output:}

\promptanswer{entailment}
\promptrelations{$\{\}$}

\textbf{Unwanted output:}

\promptrelations{\{\entails{in red clothing}{in red}\}}

\textbf{Explanation:}
The arguments of a relation may not contain prepositional phrases such as \textit{in red clothing}.

\medskip
\textbf{Example 6}

\promptinput
    {A tall man with a cap is climbing a cord.}
    {The man in a hat is climbing a rope.}

\textbf{Correct output:}

\promptanswer{entailment}
\promptrelations{$\{\entails{cap}{hat}, \entails{cord}{rope}\}$}

\textbf{Unwanted output:}

\promptrelations{$\{\entails{cap}{hat}, \entails{tall man}{man}\}$}

\textbf{Explanation:}
The relation \entails{cord}{rope} is missing.
The relation \entails{tall man}{man} is trivial because it only discards the modifier \textit{tall}.

\medskip
\textbf{Example 7}

\promptinput
    {No person is cooking.}
    {No cook is cooking in the kitchen.}

\textbf{Correct output:}

\promptanswer{entailment}
\promptrelations{$\{\entails{cook}{person}\}$}

\textbf{Unwanted output:}

\promptrelations{$\{\entails{person}{cook}\}$}

\textbf{Explanation:}
The unwanted relation does not explain the inference because negation reverses the permitted direction of lexical substitution.
It is also factually incorrect because not every person is a cook.

\medskip
\textbf{Example 8}

\promptinput
    {A person who is obese is holding a chinchilla.}
    {A fat person is holding a small animal.}

\textbf{Correct output:}

\promptanswer{entailment}
\promptrelations{$\{\entails{chinchilla}{small animal}, \entails{obese}{fat}\}$}

\textbf{Unwanted output:}

\promptrelations{$\{\entails{fat}{obese}, \entails{chinchilla}{animal}\}$}

\textbf{Explanation:}
The arguments of \entails{fat}{obese} must be reversed to align the relation with the direction of the inference.
The relation \entails{chinchilla}{animal} is insufficient because it does not account for the adjective \textit{small}, which is crucial in the hypothesis.

\medskip
\textbf{Example 9}

\promptinput
    {A little boy is laughing and happily bouncing on a trampoline outside.}
    {The child is jumping outdoors.}

\textbf{Correct output:}

\promptanswer{entailment}
\promptrelations{%
$\{\entails{little boy}{child}, \entails{bounce}{jump},$\newline
$\entails{outside}{outdoors}\}$%
}

\textbf{Unwanted output:}

\promptrelations{$\{\entails{boy}{child}, \entails{outdoors}{outside}\}$}

\textbf{Explanation:}
The relation \entails{bounce}{jump} is missing.
The relation \entails{little boy}{child} is preferred to \entails{boy}{child} because the former provides a more precise correspondence between the expressions in the premise and hypothesis.
The arguments of \entails{outdoors}{outside} must be reversed to align the relation with the direction of the inference.

\medskip
\textbf{Additional calibration while preserving all rules above:}

\begin{itemize}[leftmargin=5mm,itemsep=2pt,topsep=3pt]
    \item Only output lexical entailment relations that are both factually acceptable and needed to explain the entailment.
    \item Do not output relations for entailments that follow without a non-trivial lexical bridge.
    \item Do not use event or social implications as lexical entailment unless the phrase relation is a direct paraphrase.
    \item Do not output modifier-dropping relations such as \texttt{old woman $\rightarrow$ woman} or \texttt{military men $\rightarrow$ men}.
    \item Do not output both an inflected form and a lemmatized form for the same relation.
    \item If the best relation would violate any existing formatting rule, omit it instead of approximating it.
\end{itemize}

\medskip
Now process the following input while strictly following the above instructions and formatting.

Output exactly in one of the following formats:

\begin{quote}
\ttfamily
Answer: entailment\\
Relations: \{ \ldots \}\\
or\\
Answer: non-entailment
\end{quote}

\textbf{Input:}

\begin{quote}
\ttfamily
Premise: \${PREMISE}\\
Hypothesis: \${HYPOTHESIS}
\end{quote}

\textbf{Correct output:}

\end{promptbox}

\section{Prompt Selection}
\label{app:prompt-selection}

Preliminary prompt-selection experiments compare four controlled variants on the unanimously agreed samples (223 out of 362) using four models: %
GPT-5.4 mini, Claude Haiku 4.5, Gemini 3.1 Flash-Lite, and GPT-OSS-20B. 
Prompt variants are constructed by adding sections to the same base prompt. 
The \emph{zero-shot base} condition contains only the task definition. 
The \emph{zero-shot constrained} condition adds the relation-formatting rules from the annotation guidelines. 
The \emph{few-shot base} condition further adds nine worked examples, including correct and unwanted outputs. 
Finally, the \emph{few-shot precision} condition adds a calibration block
discouraging speculative or unnecessary relations, modifier deletion, and
duplicate surface and lemma forms.
For each prompt-model combination, we calculate multi-reference micro-F1.
We then compute the unweighted mean of the four model-level micro-F1 scores
for each prompt condition: 43.8 for \emph{zero-shot base}, 53.5 for
\emph{zero-shot constrained}, 65.2 for \emph{few-shot base}, and 69.2 for
\emph{few-shot precision}.
Based on these results, we select the \emph{few-shot precision} prompt for
the final experiments.

\section{Mixture of Figures and Tables}
\label{app:mix}

\begin{table*}[t!]
\centering
\scalebox{.8}{
\begin{tabular}{lcccccccc}
\toprule
\multirow{2}{*}{\textbf{Model}} &
\multirow{2}{*}{\textbf{Support}} &
\multirow{2}{*}{\shortstack{\textbf{Exact}\\\textbf{match}}} &
\multicolumn{4}{c}{\textbf{Micro-F1}} &
\multirow{2}{*}{\shortstack{\textbf{Micro-}\\\textbf{Precision}}} &
\multirow{2}{*}{\shortstack{\textbf{Micro-}\\\textbf{Recall}}} \\
\cmidrule(lr){4-7}
& & & \textbf{Raw $\downarrow$}
& \textbf{$\Delta$ Lem.}
& \textbf{$\Delta$ Un.}
& \textbf{$\Delta$ Lem.+Un.}
& & \\
\midrule
Gemini 3.1 Flash-Lite
& 89.0
& 65.1 {\small ($\pm$ 0.5)}
& \textbf{77.8} {\small ($\pm$ 0.6)}
& +0.4 & +0.0 & +0.4
& \textbf{70.7} {\small ($\pm$ 0.5)}
& 86.6 {\small ($\pm$ 0.7)} \\

Claude Sonnet 4.5
& 91.7
& 63.5 {\small ($\pm$ 0.2)}
& 76.9 {\small ($\pm$ 0.2)}
& +1.0 & +0.6 & +1.5
& 69.2 {\small ($\pm$ 0.2)}
& \textbf{86.7} {\small ($\pm$ 0.3)} \\

GPT-5.4
& 85.7
& \textbf{67.0} {\small ($\pm$ 0.9)}
& 76.6 {\small ($\pm$ 0.2)}
& +0.8 & +0.4 & +1.2
& 68.8 {\small ($\pm$ 0.4)}
& 86.4 {\small ($\pm$ 0.2)} \\

Gemini 3.5 Flash
& 95.2
& 61.0 {\small ($\pm$ 0.5)}
& 74.3 {\small ($\pm$ 0.5)}
& +1.1 & +0.4 & +1.5
& 66.0 {\small ($\pm$ 0.5)}
& 85.0 {\small ($\pm$ 0.5)} \\

Claude Haiku 4.5
& 89.0
& 56.8 {\small ($\pm$ 0.0)}
& 72.1 {\small ($\pm$ 0.0)}
& +1.7 & +0.0 & +1.7
& 62.2 {\small ($\pm$ 0.0)}
& 85.8 {\small ($\pm$ 0.0)} \\

\baseoracle{} WordNet \& Oracle 
& \llap{1}00.0
& 64.1 \phantom{{\small ($\pm$ 0.0)}}
& 70.5 \phantom{{\small ($\pm$ 0.0)}} 
& -- 
& -- 
& -- 
& \llap{1}00.0 \phantom{{\small ($\pm$ 0.0)}} 
& 54.4 \phantom{{\small ($\pm$ 0.0)}}\\

GPT-OSS-20B
& 77.5
& 56.7 {\small ($\pm$ 1.2)}
& 66.9 {\small ($\pm$ 0.7)}
& +5.5 & +0.2 & +5.7
& 60.1 {\small ($\pm$ 0.8)}
& 75.3 {\small ($\pm$ 0.9)} \\

GPT-5.4 Mini
& 86.9
& 44.5 {\small ($\pm$ 1.1)}
& 60.9 {\small ($\pm$ 0.8)}
& +7.9 & +0.3 & +8.4
& 50.8 {\small ($\pm$ 0.7)}
& 75.8 {\small ($\pm$ 1.0)} \\

\baseoracle{} WordNet \& Rules
& \llap{1}00.0
& 50.0 \phantom{{\small ($\pm$ 0.0)}}
& 51.0 \phantom{{\small ($\pm$ 0.0)}}
& -- 
& -- 
& -- 
& 70.2 \phantom{{\small ($\pm$ 0.0)}}
& 40.0 \phantom{{\small ($\pm$ 0.0)}}\\

Gemma 3 4B
& 92.5
& 14.9 {\small ($\pm$ 0.4)}
& 31.3 {\small ($\pm$ 0.3)}
& +11.8 & +0.1 & +12.4
& 21.9 {\small ($\pm$ 0.2)}
& 55.0 {\small ($\pm$ 0.3)} \\

\baseoracle{} Empty LEX
& \llap{1}00.0
& 37.0 \phantom{{\small ($\pm$ 0.0)}} 
& \phantom{0}0.0 \phantom{{\small ($\pm$ 0.0)}}
& -- 
& -- 
& -- 
& \llap{1}00.0 \phantom{{\small ($\pm$ 0.0)}}
& \phantom{0}0.0 \phantom{{\small ($\pm$ 0.0)}}\\
\bottomrule
\end{tabular}

}
\caption{
LLMs' results on the \LEX{}-NLI task.
These scores are more favorable to the models, as examples assigned a non-entailment label are omitted from evaluation; in contrast, \autoref{tab:lex_prediction} penalizes predictions on those examples.
Support is the percentage of problems evaluated; exclusions mostly result from models labeling problems as non-entailment.
\baseoracle{} denotes baseline and oracle systems.
}
\label{tab:lex_prediction_friendly}
\end{table*}

\begin{table*}[t]
\centering
\scalebox{.9}{
\begin{tabular}{l
    c@{\hspace{2em}}
    c@{\hspace{2em}}
    c@{\hspace{2em}}
    c}
\toprule

\multirow{2}{*}{\textbf{Model}} &
\LLMone{} \% &
\LLMa{} \%&
\LLMWN{} \%&
\LLMaWN{} \%\\

&
\dWNhead{}
&
\dWNhead{}\dgap\dAhead{}
&
\dWNhead{}
&
\dWNhead{}\dgap\dAhead{}
\\

\midrule

Claude Sonnet 4.5
& 25.4\dgap\dWN{+7.2}
& 32.0\dgap\dWN{+13.8}\dgap\dA{+6.6}
& 30.1\dgap\dWN{+11.9}
& 35.4\dgap\dWN{+17.1}\dgap\dA{+5.2} \\

Gemini 3.1 Flash-Lite
& 24.9\dgap\dWN{+6.6}
& 31.8\dgap\dWN{+13.5}\dgap\dA{+6.9}
& 29.3\dgap\dWN{+11.0}
& 34.8\dgap\dWN{+16.6}\dgap\dA{+5.5} \\

Gemini 3.5 Flash
& 23.5\dgap\dWN{+5.2}
& \textbf{33.4}\dgap\dWN{+15.2}\dgap\dA{+9.9}
& 28.2\dgap\dWN{+\phantom{1}9.9}
& \textbf{37.0}\dgap\dWN{+18.8}\dgap\dA{+8.8} \\

GPT-5.4
& 24.3\dgap\dWN{+6.1}
& 30.7\dgap\dWN{+12.4}\dgap\dA{+6.4}
& 29.3\dgap\dWN{+11.0}
& 35.6\dgap\dWN{+17.4}\dgap\dA{+6.4} \\

Claude Haiku 4.5
& \textbf{25.7}\dgap\dWN{+7.5}
& 31.2\dgap\dWN{+13.0}\dgap\dA{+5.5}
& \textbf{30.9}\dgap\dWN{+12.7}
& 35.6\dgap\dWN{+17.4}\dgap\dA{+4.7} \\

GPT-5.4 Mini
& 23.8\dgap\dWN{+5.5}
& 29.3\dgap\dWN{+11.0}\dgap\dA{+5.5}
& 29.8\dgap\dWN{+11.6}
& 34.8\dgap\dWN{+16.6}\dgap\dA{+5.0} \\

Gemma 3 4B
& 25.1\dgap\dWN{+6.9}
& 25.1\dgap\dWN{+\phantom{1}6.9}\dgap\dA{+0.0}
& 30.7\dgap\dWN{+12.4}
& 30.7\dgap\dWN{+12.4}\dgap\dA{+0.0} \\

\bottomrule
\end{tabular}
}
\caption{
LangPro proof coverage on the \textbf{test set} (of 362 entailment problems) when using various sources of the lexical KB.
LLM and \LLMa{} denote one-shot and agentic LLM-generated LEX, while \LLMWN{} and \LLMaWN{} are the former two coupled with WordNet, respectively.
Values indicate the percentage of problems proved as entailment.
The \underline{LangPro+WordNet baseline} proves 66 problems (18.2\%).
The boxed $\Delta_{\text{\textsc{wn}}}$ values indicate gains over LangPro+WordNet, while $\Delta_a$ values indicate the additional gains from agentic inference over the corresponding non-agentic LLM.
GPT-OSS-20B is omitted because the model host was unresponsive when the experiments were conducted.
}
\label{tab:langpro-362-results}
\end{table*}

\definecolor{red}{HTML}{B85C5C}
\definecolor{blue}{HTML}{557DA8}
\definecolor{green}{HTML}{5C936F}
\newcolumntype{L}[1]{>{\raggedright\arraybackslash}p{#1}}
\begin{table*}[t]
\centering
\small
\setlength{\tabcolsep}{4pt}
\renewcommand{\arraystretch}{1.15}
\scalebox{.8}{
\begin{tabularx}{1.2\textwidth}{
@{}
L{0.30\textwidth}
L{0.21\textwidth}
>{\color{red}\raggedright\arraybackslash}m{0.32\textwidth}
>{\raggedright\arraybackslash}X
@{}
}
\toprule
Premise & Hypothesis & {\color{black}Added/changed relations} & Judgment \\
\midrule
Two skiers stand near a snowy mountain. & Some humans standing. & \texttt{stand} $\sqsubseteq$ \texttt{standing}; \texttt{standing} $\sqsubseteq$ \texttt{stand} & Poor: inflection only; no new lexical knowledge \\
A black dog is running along the beach. & a dog is running on the beach & \texttt{black dog} $\sqsubseteq$ \texttt{dog}; \texttt{running along} $\sqsubseteq$ \texttt{running on}; \texttt{run along} $\sqsubseteq$ \texttt{run on}; \texttt{along} $\sqsubseteq$ \texttt{on} & Poor: modifier dropping plus invalid general substitution \\
Safety officers standing outside. & Safety officers outdoors. & \texttt{standing outside} $\sqsubseteq$ \texttt{outdoors}; \texttt{stand outside} $\sqsubseteq$ \texttt{outdoors} & Poor: semantically plausible but non-minimal \\
A female is wearing a big purple flower in her hair and has some beads around her neck. & a girl with a flower in her hair & {\color{green}\texttt{wear} $\sqsubseteq$ \texttt{with}}; {\color{green}\texttt{have} $\sqsubseteq$ \texttt{with}} & Good: unusual pairs but specific to SNLI \& SICK datasets \\
A man wearing a black hat, white shirt, red vest, black tie, and face makeup is swallowing a sword, while a man wearing a cowboy hat behind him is watching. & A man is performing. & \texttt{swallow} $\sqsubseteq$ \texttt{perform} & Poor: valid only in the staged context \\
A little girl is petting her white, fluffy cat. & The girl has a pet. & \texttt{little girl} $\sqsubseteq$ \texttt{girl}; \texttt{petting} $\sqsubseteq$ \texttt{has}; \texttt{pet} $\sqsubseteq$ \texttt{have} & Poor: modifier dropping plus false possession inference \\
Little kids are on an amusement park ride. & Children are on a ride. & \texttt{little kids} $\sqsubseteq$ \texttt{children}; \texttt{little kid} $\sqsubseteq$ \texttt{children}; \texttt{amusement park ride} $\sqsubseteq$ \texttt{ride} & Poor: unnormalized, non-minimal modifier dropping \\
Man in blue shirt playing water instrument in an orchestra. & The man is wearing blue & \texttt{in} $\sqsubseteq$ \texttt{wearing}; \texttt{in} $\sqsubseteq$ \texttt{wear} & Poor: valid only in a clothing construction \\
A man on an elevator looks at the mirror on the wall. & A person in the elevator looks at a mirror. & \texttt{on} $\sqsubseteq$ \texttt{in}; \texttt{mirror on the wall} $\sqsubseteq$ \texttt{mirror} & Poor: invalid preposition change plus modifier dropping \\
A child wearing a rainbow striped shirt is pushing a toy cart. & A child has a toy. & \texttt{push} $\sqsubseteq$ \texttt{have} & Poor: pushing does not entail possession \\
Young blond woman putting her foot into a water fountain & A person is dipping her foot into water. & \texttt{put into} $\sqsubseteq$ \texttt{dip into}; \texttt{put} $\sqsubseteq$ \texttt{dip} & Poor: valid only for the foot--water context \\
The blonde tennis player is wearing a white shirt and a green tennis skirt. & The tennis playing is wearing a green skirt. & \texttt{tennis player} $\sqsubseteq$ \texttt{tennis playing}; \texttt{tennis player} $\sqsubseteq$ \texttt{tennis play}; \texttt{player} $\sqsubseteq$ \texttt{playing} & Poor: participant--event category mismatch \\
Adjusting a tie for accuracy. & Fixing a tie. & \texttt{adjusting} $\sqsubseteq$ \texttt{fixing}; \texttt{adjusting for accuracy} $\sqsubseteq$ \texttt{fixing}; \texttt{adjust for accuracy} $\sqsubseteq$ \texttt{fix} & Poor: unnormalized and context-dependent \\
The bird's feet are grasping the window tightly. & there is a bird resting & \texttt{grasp} $\sqsubseteq$ \texttt{rest} & Poor: grasping does not entail resting \\
\bottomrule
\end{tabularx}
}
\caption{Quality judgments for the 14 additional problems obtained by Gemini 3.1 Flash-Lite in $\mathrm{LLM}^{\mathrm{WN}}_{a}$ agentic refinement over $\mathrm{LLM}^{\mathrm{WN}}_{1}$. We judge 1 case good, 0 mixed, and 13 poor.}
\label{tab:gemini-agentic-quality-plus14}
\end{table*}

\begin{table*}[t]
\centering
\small
\setlength{\tabcolsep}{4pt}
\renewcommand{\arraystretch}{1.15}
\scalebox{.8}{
\begin{tabularx}{1.2\textwidth}{
@{}
L{0.30\textwidth}
L{0.21\textwidth}
>{\color{red}\raggedright\arraybackslash}m{0.32\textwidth}
>{\raggedright\arraybackslash}X
@{}
}
\toprule
Premise & Hypothesis & {\color{black}Added/changed relations} & Judgment \\
\midrule
A black dog is walking through a stream of water. & An animal is walking through water. & {\color{green}\texttt{black dog} $\sqsubseteq$ \texttt{animal}}{\color{black}; }{\color{green}\texttt{stream} $\sqsubseteq$ \texttt{water}} & Good: both relations are generally valid; \texttt{black dog} $\sqsubseteq$ \texttt{animal} is non-minimal but sound \\
Two skiers stand near a snowy mountain. & Some humans standing. & \texttt{stand} $\sqsubseteq$ \texttt{standing}; \texttt{standing} $\sqsubseteq$ \texttt{stand} & Poor: inflection only; no new lexical knowledge \\
A black dog is running along the beach. & a dog is running on the beach & \texttt{black dog} $\sqsubseteq$ \texttt{dog}; \texttt{running along} $\sqsubseteq$ \texttt{running on}; \texttt{run along} $\sqsubseteq$ \texttt{run on}; \texttt{along} $\sqsubseteq$ \texttt{on} & Poor: modifier dropping plus invalid general substitution \\
An Asian woman in a floral wedding dress poses on a bridge close to her bridesmaids. & An asian women is wearing a wedding dress. & \texttt{woman} $\sqsubseteq$ \texttt{women}; \texttt{women} $\sqsubseteq$ \texttt{woman}; \texttt{in} $\sqsubseteq$ \texttt{wear}; \texttt{floral wedding dress} $\sqsubseteq$ \texttt{wedding dress}; \texttt{floral wed dress} $\sqsubseteq$ \texttt{wed dress} & Poor: inflection, context dependence, modifier dropping, and malformed forms \\
A lady walking down stairs with a green poster pod on her back. & A woman walking down stairs with a green thing on her back. & {\color{green}\texttt{lady} $\sqsubseteq$ \texttt{woman}} & Good: generally valid lexical entailment \\
Safety officers standing outside. & Safety officers outdoors. & \texttt{standing outside} $\sqsubseteq$ \texttt{outdoors}; \texttt{stand outside} $\sqsubseteq$ \texttt{outdoors} & Poor: semantically plausible but non-minimal \\
A man wearing a black hat, white shirt, red vest, black tie, and face makeup is swallowing a sword, while a man wearing a cowboy hat behind him is watching. & A man is performing. & \texttt{swallowing} $\sqsubseteq$ \texttt{performing}; \texttt{swallow} $\sqsubseteq$ \texttt{perform} & Poor: context-dependent and partly unlemmatized \\
A black dog runs in the grass, its tongue hanging out. & The dog is outdoors. & \texttt{run} $\sqsubseteq$ \texttt{outdoors} & Poor: running does not entail being outdoors \\
A little girl is petting her white, fluffy cat. & The girl has a pet. & \texttt{petting} $\sqsubseteq$ \texttt{having}; \texttt{pet} $\sqsubseteq$ \texttt{have} & Poor: petting does not entail possession \\
Little kids are on an amusement park ride. & Children are on a ride. & \texttt{little kids} $\sqsubseteq$ \texttt{children}; \texttt{little kid} $\sqsubseteq$ \texttt{children}; \texttt{amusement park ride} $\sqsubseteq$ \texttt{ride} & Poor: unnormalized, non-minimal modifier dropping \\
A child wearing a rainbow striped shirt is pushing a toy cart. & A child has a toy. & \texttt{push} $\sqsubseteq$ \texttt{have} & Poor: pushing does not entail possession \\
A guy kneeling down in front of a desk. & A man bending down in front of a piece of furniture. & {\color{green}\texttt{desk} $\sqsubseteq$ \texttt{furniture}} & Good: valid normalized hypernym, but not new semantic knowledge over one-shot \\
An asian lady in a brown shirt with glasses is making some food. & A woman is cooking & {\color{green}\texttt{lady} $\sqsubseteq$ \texttt{woman}}; {\color{red}\texttt{making} $\sqsubseteq$ \texttt{cooking}; \texttt{make} $\sqsubseteq$ \texttt{cook}} & Mixed: one valid relation; two food-context-dependent mappings \\
Adjusting a tie for accuracy. & Fixing a tie. & \texttt{adjusting} $\sqsubseteq$ \texttt{fixing}; \texttt{adjusting for accuracy} $\sqsubseteq$ \texttt{fixing}; \texttt{adjust for accuracy} $\sqsubseteq$ \texttt{fix} & Poor: unnormalized and context-dependent \\
The bird's feet are grasping the window tightly. & there is a bird resting & \texttt{grasp} $\sqsubseteq$ \texttt{rest} & Poor: grasping does not entail resting \\
\bottomrule
\end{tabularx}
}
\caption{Quality judgments for the 15 additional problems obtained by Gemini 3.1 Flash-Lite in $\mathrm{LLM}_{a}$ agentic refinement over $\mathrm{LLM}_{1}$. We judge 3 cases good, 1 mixed, and 11 poor.}
\label{tab:gemini-agentic-quality-plus15}
\end{table*}

\end{document}